\documentclass[10pt,twocolumn,letterpaper]{article}

\usepackage{titling}
\usepackage{cvpr}
\usepackage[dvipsnames]{xcolor,colortbl}
\usepackage{times}
\usepackage{epsfig}
\usepackage{array}
\usepackage{graphicx,booktabs}
\usepackage{amsmath}
\usepackage{amssymb}
\usepackage{arydshln}
\usepackage{subcaption}
\usepackage{multirow}
\usepackage{adjustbox}

\usepackage{cuted}
\usepackage[moderate]{savetrees}
\usepackage[pagebackref=true,breaklinks=true,letterpaper=true,colorlinks,bookmarks=false]{hyperref}

\cvprfinalcopy % *** Uncomment this line for the final submission

\def\cvprPaperID{4377} % *** Enter the CVPR Paper ID here

\ifcvprfinal\fi
\begin{document}

\newcommand{\phseo}[1]{\textcolor{red}{#1}}
\newcommand{\cs}[1]{\textcolor{blue}{#1}}

%%%%%%%%% TITLE
% \title{Multi-tasked Multimodal Video Representation Learning with \\Object-centric Co-attentional Transformation}
\title{Look Before you Speak: Visually Contextualized Utterances}

\author{Paul Hongsuck Seo~~~~~~~~~~~~~Arsha Nagrani~~~~~~~~~~~~~Cordelia Schmid\\
Google Research\\
{\tt\small \{phseo,anagrani,cordelias\}@google.com}
}

\maketitle
\noindent
\begin{strip}
    \centering\noindent
    \vspace{-1.5cm}
    \includegraphics[width=\linewidth]{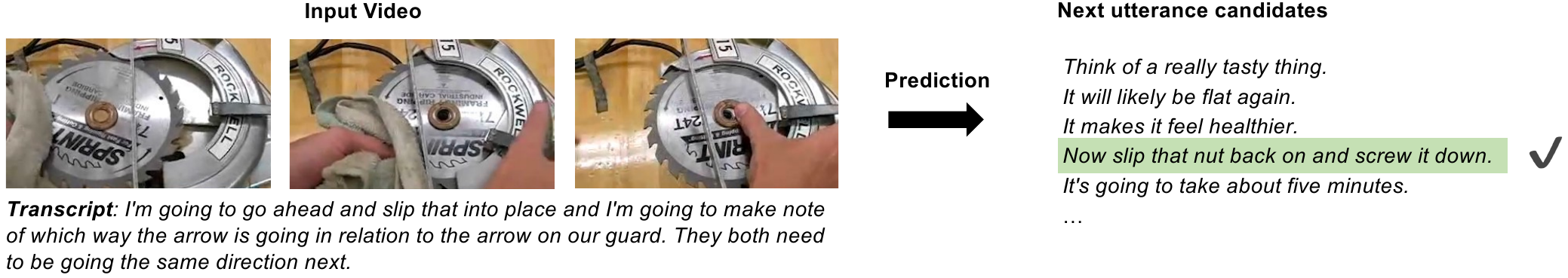}
    \vspace{-4mm}
    \captionof{figure}{\textbf{Visually Contextualised Future Utterance Prediction.} Given an instructional video with paired text and video data, we predict the next utterance in the video using a Co-attentional Multimodal Video Transformer. Our model trained on this task also achieves state-of-the-art performance on downstream VideoQA benchmarks.  }
    \label{fig:nup}
\end{strip}

%%%%%%%%% ABSTRACT
% !TEX root = ../egpaper_for_review.tex

\begin{abstract}
  While most conversational AI systems focus on \textit{textual} dialogue only, conditioning utterances on visual context (when it's available) can lead to more realistic conversations.  Unfortunately, a major challenge for incorporating visual context into conversational dialogue is the lack of large-scale labeled datasets. 
  We provide a solution in the form of a new visually conditioned Future Utterance Prediction task. Our task involves predicting the next utterance in a video, using both visual frames and transcribed speech as context. By exploiting the large number of instructional videos online, we train a model to solve this task at scale, without the need for manual annotations.
  Leveraging recent advances in multimodal learning, our model consists of a novel co-attentional multimodal video transformer, and when trained on both textual and visual context, outperforms baselines that use textual inputs alone. Further, we demonstrate that our model trained for this task on unlabelled videos achieves state-of-the-art performance on a number of downstream VideoQA benchmarks such as MSRVTT-QA, MSVD-QA, ActivityNet-QA and How2QA. 
  
  \vspace{-1em}
\end{abstract}

%%%%%%%%% BODY TEXT
% !TEX root = ../egpaper_for_review.tex

\section{Introduction}

Imagine that you are cooking an elaborate meal, but forget the next step in the recipe -- or fixing your car and uncertain about which tool to pick up next. Developing an intelligent dialogue system\footnote{often used interchangeably with the term `conversational AI'}
that not only emulates human conversation, but also
predicts and suggests future actions -- not to mention is able to answer questions on complex tasks and topics -- has long been a moonshot goal for the AI community. Conversational AI allows humans to interact with systems in free-form natural language, in the same way that we would communicate with one another. This has led to an outpouring of research in NLP focused on conversational agents, ranging from goal-oriented systems for helping with reservations~\cite{asri2017frames,williams2012belief} to chit-chat models~\cite{li2016persona,serban2015building,yan2018chitty}, both of which are found in modern virtual assistants such as Alexa, Google Assistant and Siri.

Such works, however, are limited to linguistic interactions only. In contrast, human interaction in the physical world is facilitated through multiple modalities (\eg verbal, visual, haptic), each modality often complementing the other seamlessly. While doing a task, it is often easier to \textit{show} another person your progress, than to describe it verbally. Hence we argue that a truly intelligent dialogue system would have knowledge of \textit{both} visual and textual contexts before making its next utterance. Unfortunately, a major challenge for incorporating visual context is a lack of suitable data. Most traditional conversational datasets~\cite{asri2017frames,budzianowski2018multiwoz,henderson2014second,shah2018bootstrapping} are solely text based, and notoriously difficult to collect, relying on narrowly constructed ontologies~\cite{henderson2014second,mrkvsic2015multi} and highly specific domains~\cite{asri2017frames,budzianowski2018multiwoz}. More importantly, they do not contain visual information of the surrounding physical environment. 

In an attempt to incorporate visual context to dialogue systems, the task of visual dialog~\cite{das2017visual} was proposed, which requires an AI agent to hold a meaningful dialog with humans given an image~\cite{das2017visual} or a video~\cite{kim2019eighth}. In these works, a dialog history and question are artificially created for each image/video in a dataset, and the goal is then to infer context from history and answer the question accurately. Such datasets, while valuable, are created at great manual effort and contain artificially contrived scenarios, where the dialog history is not naturally present in video.
Such datasets are also limited in size. 

Unlike such works~\cite{das2017visual,kim2019eighth}, we propose to use online videos to learn from naturally co-occurring vision and dialogue in a scalable manner.  We note that certain video domains such as \textit{narrated instructional videos}~\cite{miech2019howto100m,tang2019coin} and
\textit{lifestyle vlogs}~\cite{fouhey2018lifestyle,ignat2019identifying} are available in huge numbers
(e.g.\ online on video sharing platforms) and are likely to contain narration explicitly linked to the visual content. Given the availability of high quality ASR, this gives us a large amount of readily available paired visual and textual data. We begin by proposing a future prediction task, where the goal is to predict the next utterance in an instructional video, given both visual and textual contexts (Figure~\ref{fig:nup}). The labels for such a task are freely available from the video itself. 

As we show in this work, solving such a task requires knowledge of both visual and textual contexts. Leveraging recent advances in multimodal learning, we do so with a two-stream co-attentional transformer based model, each stream encoding a different modality. 
Our co-attentional model effectively attends to features within each modality, as well as across modalities through lateral self-attention blocks. We demonstrate that using both visual and textual information leads to a large performance gain over using text alone, and additionally, our two-stream co-attentional model outperforms single stream multimodal models. 
In addition, we show that our model, trained for this future prediction task, can be transferred to other conversational tasks, achieving state-of-the-art performance on various VideoQA benchmarks.

Concretely, we make the following four contributions: (i) We formulate a future utterance prediction (FUP) task which uses \textit{both} dialogue and vision; (ii) We re-purpose freely available online instructional video datasets to create training and testing benchmarks for this task (HowToFUP); (iii) We propose a new two-stream multimodal video transformer based architecture (CoMVT) which effectively attends jointly over words in text and visual objects and scenes to learn visual-dialogue context; and finally (iv)~We show that our model trained on unlabelled instructional videos is also, perhaps surprisingly, able to achieve state-of-the-art performance on a number of downstream vision-language QA datasets, including MSRVTT-QA~\cite{xu2017video}, MSVD-QA~\cite{xu2017video}, ActivityNet-QA~\cite{yu2019activitynet}, and How2QA~\cite{li2020hero}. 

% !TEX root = ../egpaper_for_review.tex

\section{Related Work}
\label{sec:related_work}
\noindent\textbf{Future Utterance Prediction.}
Predicting future utterances from textual data alone has been widely explored in the NLP community for conversational AI systems.
Approaches include hand-crafted rules~\cite{deemter2005real,varges2001instance}, example-based agents~\cite{kim2007example,lee2006situation} and modern neural networks~\cite{serban2016building,shao2017generating}, and aim to generate realistic responses for goal-oriented dialog systems or chatbots. Future prediction has also been used as an unsupervised pretraining objective for text corpora, \eg next sentence prediction in BERT~\cite{devlin2018bert}.
Unlike these works, we focus on jointly learning from visual context as well as text. Related to our work is the task of scene-aware dialog prediction~\cite{alamri2019audio,hori2018audio}, where the goal is to answer questions grounded to a video clip input, given a manually created dialog history. A number of works show promising results on this task~\cite{chu2020multi,hori2019end,le2020multimodal,lee2020dstc8,li2020bridging}, however they rely on manually created VQA datasets.
\noindent\textbf{Vision and Language Tasks.}
Popular vision and language tasks include visual question answering~\cite{anderson2018bottom,antol2015vqa, fukui2016multimodal,goyal2017making,mun2017marioqa,noh2016image}, visual dialog~\cite{das2017visual,das2017learning,jin2019video,kottur2018visual,seo2017visual}, visual captioning~\cite{mun2017text,mun2019streamlined,seo2020reinforcing,sharma2018conceptual,xu2015show}, visual grounding~\cite{deng2018visual,fukui2016multimodal,mun2020local,xiao2017weakly} and video-text retrieval~\cite{bain2020condensed,gabeur2020Learning,liu2019use,miech2018learning,patrick2020support}. 
There have also been attempts to use transcribed speech in videos as a source of weak supervision~\cite{miech2020end,nagrani2020speech2action, sun2019learning, sun2019videobert}, where the goal is to learn a good visual encoder, and consequently such works are largely evaluated only on  downstream tasks that involve unimodal video frame inputs. In contrast, we learn an encoder that can effectively learn to \textit{co-attend} to both vision and text, and is useful for downstream tasks that involve both modalities. 

\noindent\textbf{Multimodal Vision-Text Architectures.}
A large number of multimodal architectures focus on late fusion of modalities, with popular choices being summation, concatenation and canonical correlation analysis~\cite{ben2018deep,sharma2020deep,sun2020Learning}.
\cite{li2020hero} encodes multimodal inputs in a hierarchical structure, where the local context of a video frame is captured by a Cross-modal Transformer via multimodal fusion, and global video context is captured by a Temporal Transformer. Here cross-modal interactions are limited to a single segment only (where the input modalities are aligned), covering a short timespan. This does not allow multimodal interactions between non-aligned inputs - and we note that the content of human speech is not always precisely aligned with its corresponding visual contexts in time~\cite{miech2020end}. In contrast, our method allows global cross-modal interactions, unconstrained by temporal alignment.
More recent works~\cite{gabeur2020Learning,li2020bridging} explore deeper interactions between video frame features and features from other modalities, by fusing modalities earlier, at the input level itself.
In these works, however, inputs from multiple modalities are fed into a single transformer, with `modality specific' encodings to distinguish between the modalities. 
In contrast, our two stream transformer decouples within-modality interactions in individual modality streams and allows cross-modality interactions with lateral self-attention blocks.
Another differentiater is the fact that all these works operate on scene-level features, while we focus on objects. Thanks to off-the-shelf high-quality object detection \cite{han2019re,redmon2016you} and simple bounding box representations, object-centric features have been widely used in a number of image and language problems~\cite{anderson2018bottom,lu2018neural}. In particular, ViLBERT~\cite{lu2019vilbert} feeds object features and language inputs to a co-attentional transformer. Extending such methods to video, however, is  non-trivial, given the number of frames in a video. While object detectors work well on single frames, obtaining reliable tracklet based video features is still an open problem. 

% !TEX root = ../egpaper_for_review.tex

\begin{figure*}
    \centering
    \includegraphics[width=\linewidth]{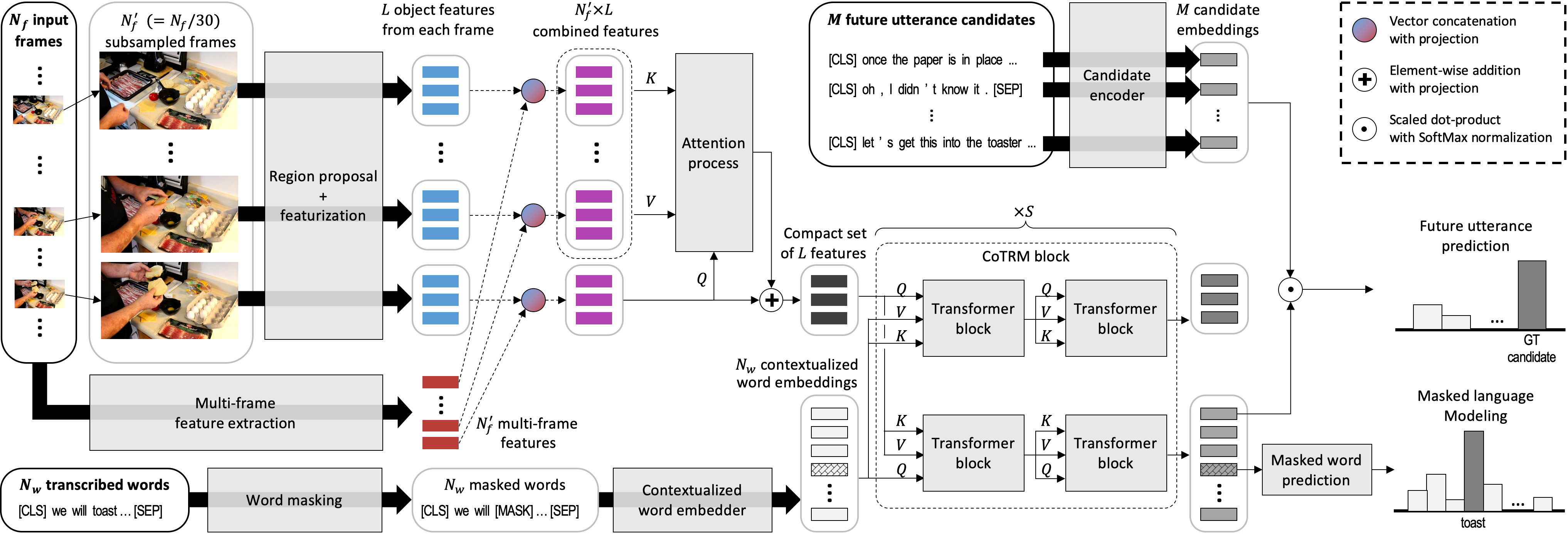}
    \caption{\textbf{Co-attentional Multimodal Video Transformer:} Visual depiction of our multimodal training network with $L=3$ object features per frame. Our model consists of 2 streams, a visual stream which operates on spatiotemporal features, and a text steam which ingests word level features. Our model is trained using two losses, a future utterance prediction ranking loss and a masked language modelling loss.}
    \label{fig:architecture}
\end{figure*}

\section{Future Utterance Prediction}
\label{sec:tasks}

We begin by proposing a new future utterance prediction task. The goal is to predict the next utterance in a video, given the previous multimodal context (Figure~\ref{fig:nup}).
While next utterance prediction can be evaluated as a generative task~\cite{das2017visual,gao2018neural,lee2020dstc8}, we simplify the problem to be selection among pre-collected candidates.
Precisely speaking, given a video clip $\mathcal{V}=\left(F,W\right)$ where $F=\left\{f_i\right\}_{i=1}^{N_f}$ is a sequence of video frames and $W=\left\{w_i\right\}_{i=1}^{N_w}$ is a sequence of transcribed words, our goal is to select the true next utterance $u_\mathcal{T}$ from a set of candidates $U=\left\{u_i\right\}_{i=1}^{M}$ where $\mathcal{T}$ is the index of the true element in $U$ and we set $M=100$ (Figure~\ref{fig:nup}).
Performance is then assessed by ranking the candidates and using popular retrieval metrics.  

Our reasons for opting for ranking rather than generation are two fold: (i) Metrics for evaluating language generation (\eg \ BLEU~\cite{papineni2002bleu}, METEOR~\cite{banerjee2005meteor}) are focused on local matches (n-grams, longest matching sequences, etc). By definition, such metrics are limited to local contexts, and struggle to account for complex sentence structures and word semantics. It is therefore widely accepted that these metrics do not align well with human ratings~\cite{seo2020reinforcing}; and (ii) The output distribution in sentence generation is multimodal, \ie the same information can be paraphrased in multiple ways, leading to many correct answers. There is also inherent ambiguity in the future -- given the observation of the present, multiple predictions
about the future are possible~\cite{furnari2018leveraging}. In language generation tasks, this is handled by collecting multiple ground truth answers, a strategy which is expensive and difficult to scale. In contrast, popular ranking metrics such as recall@$k$ are better able to assess model performance in tasks with a multimodal output distribution~\cite{das2017visual,kummerfeld2019large}.

We note here that unlabelled videos can be used to generate data for our task in a scalable manner. A list of future candidate utterances can be created automatically, with the positive sample being the next utterance in the video, and making the assumption that randomly sampled utterances from different video clips are likely to be negatives~\cite{sun2019learning}. 

In this work, we use the videos from the HowTo100M dataset~\cite{miech2019howto100m}, as this is a large dataset of 1.2M instructional videos where the speech is usually explicitly linked to the visual content in the video. Examples of future prediction candidates for this dataset can be seen in Figure~\ref{fig:qualitative}. % and in the supplementary material. 
We use 90\% of the videos in HowTo100M for training, and reserve 5\% each for validation and test respectively.
We name this benchmark How2FUP (more details are provided in Section~\ref{sec:datasets}).

% !TEX root = ../egpaper_for_review.tex

\section{Model}
Our goal is to effectively learn from both vision and text in a video. We propose  a Co-attentional Multimodal Video Transformer (CoMVT), which 
given a video clip $\mathcal{V}$, extracts contextualized word embeddings and visual features from transcribed words and video frames respectively, and fuses the extracted features to form a multimodal video feature using a co-attentional transformer.
We first describe our network architecture, and then the losses used to train this model to solve the task described above (both network and losses are visually depicted in Figure~\ref{fig:architecture}). 
\label{sec:methods}
\subsection{Network Architecture}
\subsubsection{Text Input Features}
Given a sequence of transcribed words\footnote{We use WordPiece tokenization from the BERT vocabulary.} $W$, 
we extract $N_w=|W|$ contextualized word embeddings $e_i$ using BERT~\cite{devlin2018bert}. 

\subsubsection{Visual Input Features}
 We extract two types of visual features -- multi-frame scene level features, and object level features extracted per frame. 

\noindent\textbf{Scene-level:} We first extract $N_f'$ multi-frame features $m_i$ by feeding $F=\left\{f_i\right\}_{i=1}^{N_f}$ frames into S3D~\cite{xie2018rethinking}, a 3D CNN model which has been used in previous approaches in multimodal representation learning~\cite{sun2019learning,sun2019videobert}.
Note that $N_f' (\le N_f)$ is the number of extracted multi-frame features (determined by the stride and temporal downsampling rate of S3D).
Similarly to~\cite{sun2019videobert}, for every non-overlapping 1-second-long segment of the video, we sample 30 frames and obtain a single feature vector by applying global average-pooling spatiotemporally to the feature activations before the final classifier. This gives us one multi-frame feature $m_i$ per second.

\noindent\textbf{Object-level:} While multi-frame features are effective at capturing spatiotemporal dynamics, they limit access to individual concepts or objects by squeezing information into a single vector.
To overcome this, we also extract object-level features corresponding to single visual objects in each frame.
We first subsample $N_f'$ frames building $F'=\left\{f_i'\right\}_{i=1}^{N_f'}$ where each frame $f_i'$ is temporally aligned with a multi-frame feature $m_i$.
Single-frame object features are then extracted from top-scoring bounding box proposals in each $f_i'$.
Following \cite{changpinyo2019decoupled}, bounding boxes are proposed by a region proposal network (RPN) in \cite{ren2015faster} and featurized using Graph-Regularized Image Semantic Embeddings (Graph-RISE)~\cite{juan2019graph}.
That is, the single-frame object features $\left\{o_{ij}\right\}_{j=1}^{L}$ are extracted from each frame $f_i'$ by
\vspace{0.1cm}
\begin{align}
    B_i &= \mathrm{RPN}(f_i') \\
    o_{ij} &= \mathrm{Graph\text{-}RISE}(b_{ij};f_i'),\hspace{0.3cm} b_{ij} \in B_i
\end{align}
\vspace{-0.2cm}

\noindent where $B_i=\left\{b_{ij}\right\}_{j=1}^{L}$ is a set of top $L$ bounding box in $f_i'$ proposed by RPN.

Note that $o_{ij}$ is object-specific but lacks temporal information whereas $m_i$ encodes temporal dynamics without allowing object-specific access.
Therefore, we construct combined spatiotemporal visual features that provide both temporal information and object-specific access by merging these two types of features:
\vspace{0.1cm}
\begin{align}
    %\hat{o}_{ij} = g_\mathrm{comb}([o_{ij};m_i])+\mathrm{pos}(o_{ij}) \label{eq:sp-obj-feat}
    v^\mathrm{st}_{ij} = g_\mathrm{comb}([o_{ij};m_i])+\mathrm{pos}(o_{ij}) \label{eq:sp-obj-feat}
\end{align}
\vspace{-0.2cm}

\noindent where $[;]$ and $g_\mathrm{comb}$ are a vector concatenation operation and a two-layer perceptron, respectively, and $\mathrm{pos}(o_{ij})$ is a positional encoding.
The positional encoding $\mathrm{pos}(o_{ij})$  captures the spatiotemporal location of an object $o_{ij}$ and is computed by the sum of sinusoidal encoding of temporal location~\cite{vaswani2017attention} and linear projection of bounding box coordinates\footnote{We use the normalized coordinates of top-left and bottom-right corners of bounding boxes.}.

\noindent\textbf{Compact feature set extraction:}
% \paragraph{Compact feature set extraction:}
Since the set of combined visual features $\{v^\mathrm{st}_{ij}\}$ is extracted from the entire spatiotemporal space of a video, this leads to a large number of features (often redudant as the same object is detected in multiple frames), which significantly increases complexity of the transformer (see Table \ref{tab:compact_features}).
Therefore, we construct a more compact set of visual features by aggregating redundant features through an attention process. %(Figure~\ref{fig:architecture}).
We select a temporal anchor point $t$ where the object features $v^\mathrm{st}_{tj}$ in frame $f_t'$ are used as queries. We then attend to the remaining features as follows: $V_\mathrm{target} = \left\{v^\mathrm{st}_{ij'}|i \ne t\right\}$.
This way, the model retrieves features relevant to the query (learned during training).
Formally, our model computes a scalar attention score $\alpha_j(v)$ for each element $v\in O_\mathrm{target}$ from a $j_\text{th}$ query object, and obtains an attended visual feature $v^\mathrm{att}_j$ by
\vspace{0.1cm}
\begin{align}
    \alpha_j(v) &= \frac{g_\mathrm{query}(v^\mathrm{st}_{tj}) \cdot g_\mathrm{key}(v)}{\sum_{v'\in V_\mathrm{target}}g_\mathrm{query}(v^\mathrm{st}_{tj}) \cdot g_\mathrm{key}(v')}  \\
    v^\mathrm{att}_j &=g_\mathrm{output}\left(\sum_{v'\in V_\mathrm{target}}\alpha_j(v') g_\mathrm{value}(v')\right) 
\end{align}
\vspace{-0.1cm}

\noindent where $g_\mathrm{query}$, $g_\mathrm{key}$, $g_\mathrm{value}$ and $g_\mathrm{output}$ are all linear projection functions, and $\cdot$ is a scaled dot product proposed in \cite{vaswani2017attention}.
The final visual feature $v^\mathrm{compact}_j$ is then computed from the sum of the anchor object features $v^\mathrm{st}_{tj}$ and the attended object features $v_j^\mathrm{att}$:
\vspace{0.1cm}
\begin{equation}
    v^\mathrm{compact}_j = g_\mathrm{proj}(v^\mathrm{st}_{tj}+v^\mathrm{att}_j)
\end{equation}
where $g_\mathrm{proj}$ is a projection function implemented by a two-layer perceptron.

\subsubsection{Co-attentional Transformers}
Features from text $w_i$ and vision $v^\mathrm{compact}_j$ are then fused using an architecture similar to the co-attentional transformer (CoTRM) proposed in \cite{lu2019vilbert}.
A CoTRM block is composed of four transformer (TRM) blocks that compute attention distributions over values by computing a scaled dot product between queries and keys, and obtaining a set of output vectors from weight-averaged values~\cite{vaswani2017attention}. 

A CoTRM block consists of two streams, each built by stacking two TRM blocks. The first TRM block in each stream takes two multimodal inputs sets: one for queries and the other for keys and values alternating their roles in each stream. The second TRM block is independant within a modality stream.

Formally, given two sets of input features $V^{(s)}$ and $E^{(s)}$, the visual features in $V^{(s)}$ at $s_\text{th}$ CoTRM block are contextualized by
\vspace{0.1cm}
\begin{align}
    \hat{V}^{(s)} &= \mathrm{TRM}(V^{(s)}, E^{(s)}) \label{eq:co-trm}\\
    V^{(s+1)} &= \mathrm{TRM}(\hat{V}^{(s)},\hat{V}^{(s)}) \label{eq:trm}
\end{align}
\vspace{-0.2cm}

\noindent where $\mathrm{TRM}(Q,K)$ is a TRM block with query inputs $Q$ and key-value inputs $K$.
Note that the first TRM block, \ie equation~\eqref{eq:co-trm}, performs inter-modality contextualization by adding related word features to each visual feature whereas the second block, \ie equation~\eqref{eq:trm}, performs intra-modality contextualization through the regular self-attention mechanism.
Similarly, the word embeddings $E^{(s)}$ are contextualized by 
\vspace{0.1cm}
\begin{align}
    \hat{E}^{(s)} &= \mathrm{TRM}(E^{(s)}, V^{(s)}) \\
    E^{(s+1)} &= \mathrm{TRM}(\hat{E}^{(s)},\hat{E}^{(s)}).
\end{align}
\vspace{-0.2cm}

\noindent We repeat this process $S$ times and set the initial inputs
$V^{(0)}= \left\{v^\mathrm{compact}_i\right\}_{i=1}^{L}$ and $E^{(0)}=\left\{e_i\right\}_{i=1}^{N_w}$.
% $V^{(0)}=\left\{v_*\right\} \bigcup \left\{v_i\right\}_{i=1}^{L}$ and $E^{(0)}=\left\{e_*\right\} \bigcup\left\{e_i\right\}_{i=1}^{N_w}$ where $v_*$ and $e_*$ are optional vectors used to encode the entire information in each modality.
% For the visual modality, we set $v_*=\frac{1}{L}\sum_{i=1}^Lv_i$ and add it only for the action anticipation task whereas $e_*$ correspond to the embedding of the special [CLS] token in the textual modality.

The two stream nature of CoTRMs inherently treats each modality separately allowing modality-specific operations and representations through different parameterizations of TRMs in the streams.

\subsection{Training Objectives}
We train our model with the following two losses: \\
\noindent\textbf{1) Next Utterance Prediction Loss:}
Here we treat the textual modality as the main modality and treat $e^{(S)}_1$ as the embedding of all multimodal inputs.
Note that $e^{(S)}_1$ corresponds to the contextualized embedding of the special `[CLS]' token added to the input.
Since our goal is to choose the true next utterance $u_\mathcal{T}$ from a set of candidate utterances $U=\left\{u_i\right\}_{i=1}^{M}$, we first embed each candidate utterance using an additional BERT encoder and predict the probability of $u_i$ being the true next utterance $P(u_i|e^{(S)}_1,U)$ by
\vspace{0.1cm}
\begin{equation}
    P(u_i|e^{(S)}_1,U) = \frac{\mathrm{exp}\left(e^{(S)}_1\cdot g_\mathrm{cand}(u_i)\right)}{\sum_{u'\in U}\mathrm{exp}\left(e^{(S)}_1\cdot g_\mathrm{cand}(u')\right)}
\end{equation}
\vspace{-0.1cm}

\noindent where $g_\mathrm{cand}$ is the candidate embedding function for which we use a BERT encoder.
Note that this candidate utterance encoder is also fine-tuned during training and has different parameters from those in the input text encoder.
We train the network by minimizing the negative log-likelihood of the true next utterance $-\mathrm{log}P(u_\mathcal{T}|e^{(S)}_1,U)$
where $U$ is constructed by collecting all ground-truth next utterances of examples within each batch during training.

\noindent\textbf{2) Masked Language Modelling:}
In addition to our next utterance prediction loss, we implement the masking scheme and loss function introduced in \cite{devlin2018bert} and mask out some of input words. %As detailed in Sec. ~\ref{}, 
We apply this loss on downstream evaluations as well. This appears to have a regularisation effect, similar to dropout~\cite{srivastava2014dropout}. 
We additionally explored visual input masking as in \cite{sun2019learning}, but found little changes in performance. 

\noindent\textbf{Implementation details.}
We use the BERT base model for the contextualized word embedding extraction and test the proposed networks with $S\in\left\{1,2,4\right\}$.
RPN~\cite{ren2015faster}, GRAPH-Rise~\cite{juan2019graph} and S3D~\cite{xie2018rethinking} are initialized and fixed with pretrained weights; all the other parameters are updated during training in all experiments.
We set the maximum lengths for the transcribed words $N_w$ and the downsampled video frames $N_f'$ to 128 and 30, respectively, and truncate longer sequences keeping the last elements.

% !TEX root = ../egpaper_for_review.tex

\section{Experiments}
\label{sec:experiments}
We first train our model for Future Utterance Prediction and show results on two datasets, HowToFUP and Coin-FUP.  
We then take the model pretrained on this task and demonstrate that it generalises well to VideoQA datasets, achieving state-of-the-art results. 
The input/output configurations for these tasks are described in Appendix~\ref{sec:tasks_app}.
The next section describes all the datasets used in this work, and then delves into experimental details. 

\subsection{Datasets}
\label{sec:datasets}
\subsubsection{Future Utterance Prediction}
\noindent\textbf{HowToFUP:}
We repurpose HowTo100M~\cite{miech2019howto100m}, a large-scale dataset of 1.2M instructional videos for the task of Future Utterance Prediction. Transcripts are obtained using the YouTube ASR API~\cite{youtubeapi}, however these are noisy (Figure~\ref{fig:qualitative_1} in the Appendix shows an example). 
Videos that have been taken down from YouTube are not used. We then divide these videos into shorter segments, henceforth referred to as video clips. The duration of video clips is determined as follows: we start with a single ASR sentence and then iteratively expand the length of the video clip backwards by adding previous sentences until the segment is longer than 5 seconds. Each video clip therefore contains full sentences in the ASR (no sentences are cut-off mid way). This process results in 35M training examples and 2M examples each in the validation and test splits. In order to create diverse validation and test sets, we then further reduce the number of clips in each by randomly subsampling 6\% of the clips (as many clips contain redundant input contexts).
The final validation and test splits consist of 120K clips, and are used for testing all models.

For each video clip, we then create a list of $M=100$ future utterance candidates through random sampling. 
$M-1$ negative candidates are sampled from the entire answer pool to build $U$ for the test and validation splits. 

We note that this dataset is an order of magnitude larger in number of datapoints than existing video captioning datasets, as well as Conceptual Captions~\cite{sharma2018conceptual}, the largest publicly released image captioning dataset widely used for pretraining vision-text models in the image domain~\cite{chen2019uniter,lu2019vilbert}, however is noisier due to (i) errors caused by imperfect ASR and (ii) given the ASR is generated from continuous narration, it often consists of incomplete sentences that lack punctuation. A further analysis in provided in Appendix~\ref{sec:asrerror}. \\ 
\noindent\textbf{COIN-FUP:} 
We also repurpose COIN~\cite{tang2019coin}, another dataset of instructional videos to evalute the task of future utterance prediction. This dataset is smaller, with 12K videos.
We follow the same clip generation pipeline used for HowToFUP\footnote{However we do not subsample in the validation and test sets, to maintain a reliable number of samples for evaluation.} to create COIN-FUP, resulting in 78K, 5K and 4K examples in the train, validation and test splits respectively.

\vspace{-0.3cm}
\subsubsection{Next Step Prediction}
\noindent\textbf{COIN-NSP:} 
Unlike HowTo100M, COIN~\cite{tang2019coin} also contains additional manually annotated categorical steps labelled for each video. Hence we also investigate the performance of a related, albeit slightly different task on this dataset -- next step prediction (NSP). Unlike FUP, where the goal is to select from a list of utterances in free form natural language, NSP focuses on predicting the next step from a list of  pre-defined action categories. 
Similarly to FUP example generation, we automatically construct COIN-NSP by iterating over each step annotation and extract its precedent multimodal video segment generating 18K training examples and 1K examples for both validation and test splits with 735 step classes (24.5 examples per class on average in the entire dataset). Note that COIN-NSP is smaller than COIN-FUP in the number of examples since there are fewer manual step annotations than total number of utterances. 
At inference time, we simply replace the candidate selection component in our model with a softmax classifier.  \\
\noindent\textbf{CrossTask-NSP:} 
CrossTask~\cite{zhukov2019cross} is another dataset that contains manual annotations of steps for instructional videos of 18 pre-selected tasks.
We create CrossTask-NSP following the COIN-NSP construction process resulting in 14K training examples and 2K validation/test examples with 105 target classes.

\vspace{-0.3cm}
\subsubsection{Downstream VideoQA Benchmarks}
\noindent\textbf{MSRVTT-QA and MSVD-QA:}
MSRVTT-QA and MSVD-QA are popular video question answering benchmarks introduced in \cite{xu2017video}.
%extending MSR-VTT~\cite{xu2016msr} and MSVD~\cite{chen2011collecting}, respectively.
We use publicly available features and follow the standard train, val and test splits used in~\cite{xu2017video}: 158K, 12K and 73K QA pairs for MSRVTT-QA and 31K, 6K and 13K pairs for MSVD-QA.

{\noindent\textbf{ActivityNet-QA:}
ActivityNet-QA~\cite{yu2019activitynet} contains 58K open-ended QA annotations where the train, val and test splits have 32K, 18K and 8K QA pairs, respectively.}

\noindent\textbf{How2QA:}
How2QA~\cite{li2020hero} consists of QA annotations for the HowTo100M dataset. It contains 35K train and 3K publicly available val samples. Each question has three negative answers and one correct answer.

\begin{figure*}[t]
    \centering
    \scalebox{0.8}{
        \begin{tabular}{cc}
            \toprule
            \textbf{Inputs (video frames and utterances)} & \textbf{Prediction (future utterance)} \\ \midrule
            
            % Example1: -humHGoKXzw-00053
            %   YT id: -humHGoKXzw
            %   start_time: 189.359s
            %   end_time:   195.930s
            \begin{tabular}{@{}p{13.3cm}@{}}
                \centering
                \adjincludegraphics[valign=M,width=0.325\linewidth]{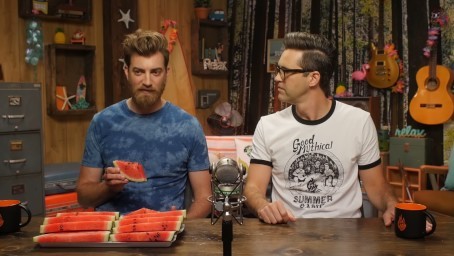}
                \adjincludegraphics[valign=M,width=0.325\linewidth]{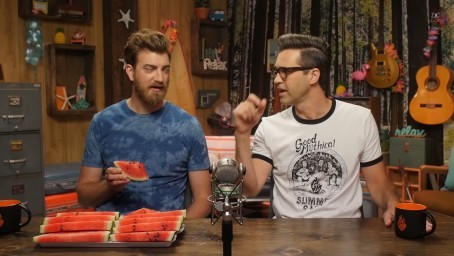}
                \adjincludegraphics[valign=M,width=0.325\linewidth]{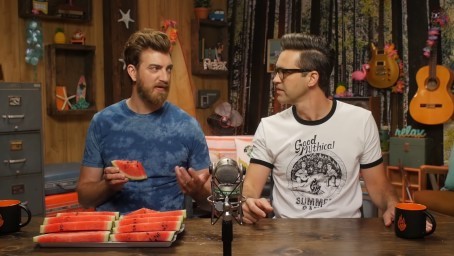} \tabularnewline[1.15cm]
                \textbf{Transcript}: It was completely rotten on the inside rotten. Yeah, because I had you waited till a because I thought it's gonna keep getting bigger, but it was just because I didn't have a green thumb.  
            \end{tabular}
            & 
            \bgroup
            \def\arraystretch{1.3}
            \begin{tabular}{@{}p{6.5cm}c@{}}
                \cellcolor[rgb]{0.9784,0.7725,0.7059}
                \raggedright \textit{It's grown.} \tabularnewline
                \raggedright \textit{Now do we have a finished product?}\tabularnewline
                \cellcolor[rgb]{0.7725,0.8784,0.7059}
                \raggedright \textit{That was my \textbf{watermelon} had completely ripened weeks ago.} &
                \adjincludegraphics[valign=M,width=0.025\linewidth]{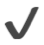}
                \tabularnewline
                \raggedright \textit{I don't think I would drive a bit long enough to get in there.}\tabularnewline
                \raggedright $\boldsymbol{\cdots}$\tabularnewline
            \end{tabular}
            \egroup \\
            \midrule

            % Example1: -Oxdgsewk5s-00076
            %   YT id: -Oxdgsewk5s
            %   start_time: 327.110s
            %   end_time:   343.130s
            \begin{tabular}{@{}p{13.3cm}@{}}
                \centering
                \adjincludegraphics[valign=M,width=0.325\linewidth]{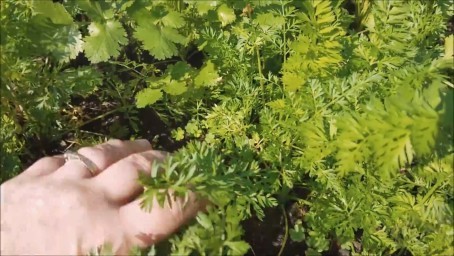}
                \adjincludegraphics[valign=M,width=0.325\linewidth]{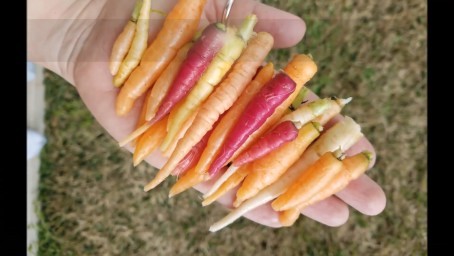}
                \adjincludegraphics[valign=M,width=0.325\linewidth]{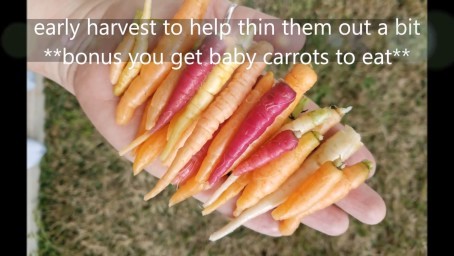} \tabularnewline[1.15cm]
                \textbf{Transcript}: And here's another one that might be close, but I just don't think they're quite ready yet, but it's always nice to check so they need a little bit more growth time. All right. 
            \end{tabular}
            & 
            \bgroup
            \def\arraystretch{1.3}
            \begin{tabular}{@{}p{6.5cm}c@{}}
                \raggedright \textit{We like to eat with bone like quit bone.}\tabularnewline
                \raggedright \textit{You need to see what these things are.}\tabularnewline
                \cellcolor[rgb]{0.7725,0.8784,0.7059}
                \raggedright \textit{So here are the \textbf{carrots}.} &
                \adjincludegraphics[valign=B,width=0.025\linewidth]{figures/qualitative_examples/checkmark.png}
                \tabularnewline
                \cellcolor[rgb]{0.9784,0.7725,0.7059}
                \raggedright \textit{They're looking fabulous.} \tabularnewline
                \raggedright $\boldsymbol{\cdots}$\tabularnewline
            \end{tabular}
            \egroup \\
            \midrule

            % Example1: -GfL5avNKEY-00014
            %   YT id: -GfL5avNKEY
            %   start_time: 100.650s
            %   end_time:   106.439s
            \begin{tabular}{@{}p{13.3cm}@{}}
                \centering
                \adjincludegraphics[valign=M,width=0.325\linewidth]{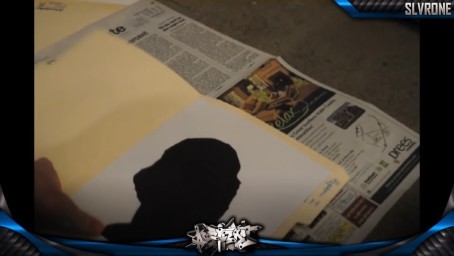}
                \adjincludegraphics[valign=M,width=0.325\linewidth]{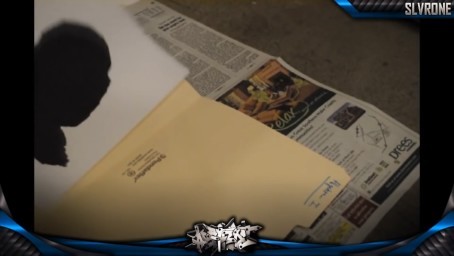}
                \adjincludegraphics[valign=M,width=0.325\linewidth]{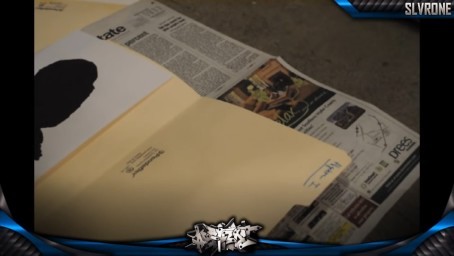} \tabularnewline[1.15cm]
                \textbf{Transcript}: They just shake it up and pretty much the exact same thing as spraying you all should be familiar with heavy scraping.
            \end{tabular}
            & 
            \bgroup
            \def\arraystretch{1.3}
            \begin{tabular}{@{}p{6.5cm}c@{}}
                \raggedright \textit{That's bad.}\tabularnewline
                \cellcolor[rgb]{0.7725,0.8784,0.7059}
                \raggedright \textit{But so basically I just spray the \textbf{folder} and then I just paste this down and I try to get them in all the exact same position.} &
                \adjincludegraphics[valign=T,width=0.025\linewidth]{figures/qualitative_examples/checkmark.png}
                \tabularnewline
                \cellcolor[rgb]{0.9784,0.7725,0.7059}
                \raggedright \textit{It doesn't take much time at all.} \tabularnewline
                \raggedright \textit{We need to mount this on our a/c condenser.}\tabularnewline
                \raggedright $\boldsymbol{\cdots}$\tabularnewline
            \end{tabular}
            \egroup \\
            \midrule
            
            % Example1: 05bT5dz8e3M-00031
            %   YT id: 05bT5dz8e3M
            %   start_time: 157.180s
            %   end_time:   164.030s
            \begin{tabular}{@{}p{13.3cm}@{}}
                \centering
                \adjincludegraphics[valign=M,width=0.325\linewidth]{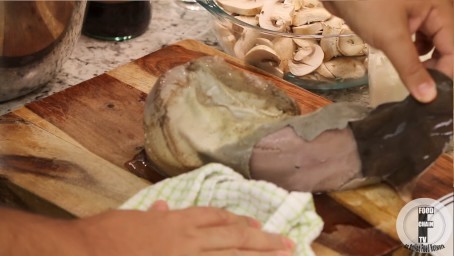}
                \adjincludegraphics[valign=M,width=0.325\linewidth]{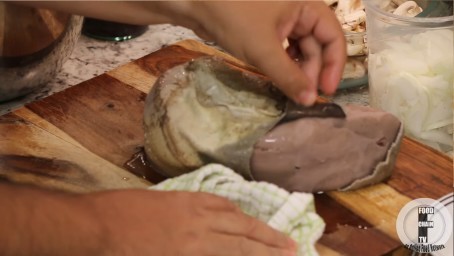}
                \adjincludegraphics[valign=M,width=0.325\linewidth]{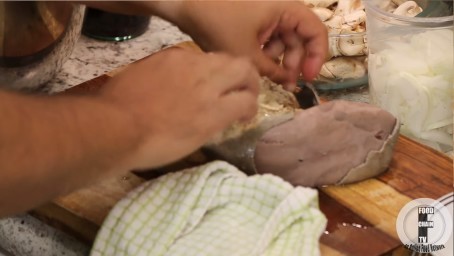} \tabularnewline[1.15cm]
                \textbf{Transcript}: This skin should come right off just like that. Okay, and there we go.
            \end{tabular}
            & 
            \bgroup
            \def\arraystretch{1.3}
            \begin{tabular}{@{}p{6.5cm}c@{}}
                \cellcolor[rgb]{0.7725,0.8784,0.7059}
                \raggedright \textit{So underneath that is some really really nice \textbf{meat}.} &
                \adjincludegraphics[valign=M,width=0.025\linewidth]{figures/qualitative_examples/checkmark.png}
                \tabularnewline
                \raggedright \textit{It is nice to write down a date on the jar.}\tabularnewline
                \cellcolor[rgb]{0.9784,0.7725,0.7059}
                \raggedright \textit{I guess the skin anyway, so it's ready.} \tabularnewline
                \raggedright \textit{Don't click it.}\tabularnewline
                \raggedright $\boldsymbol{\cdots}$\tabularnewline
            \end{tabular}
            \egroup \\
            \bottomrule

        \end{tabular}
    }
    \caption{
    \textbf{Qualitative results on HowToFUP.} 
    On the right, we show the results of the baseline model that uses text inputs only (highlighted in \colorbox[rgb]{0.9784,0.7725,0.7059}{red}) and our multimodal model (highlighted in \colorbox[rgb]{0.7725,0.8784,0.7059}{green}). The GT utterance has a $\checkmark$ next to it. 
    Note how the transcript often contains phrases with subtle indications to visual content, such as `here's another one' (second row) and `should come off right like that' (fourth row). In many of these cases, the correct future utterance refers to an object which can only be known from the visual context (highlighted in bold). The text only model often selects generics utterances, or those which are referred to specifically in previous dialogue (fourth row, selected candidate has the word `skin'). 
    Further examples are provided in Appendix~\ref{sec:qual}.
    }
    \label{fig:qualitative}
\end{figure*}

\subsection{Baselines}
\label{sec:baseline}
% S3D need to be explained
We compare our model to a number of single and multiple modality baselines. \\
\noindent\textbf{S3D - visual only:} We use the S3D~\cite{xie2018rethinking} model pretrained on Kinetics applied to video frames. \\
\noindent\textbf{Text only Baseline:} For the text only baseline, we use BERT~\cite{devlin2018bert}, which is the winning model in the Eighth Dialog System Technology Challenge for response prediction in text only dialog benchmarks~\cite{ws-aaai-dstc20task2}. \\
\noindent\textbf{Single-Stream Multimodal Baseline:} We also implement a single stream transformer operating on a single multimodal input stream, which is the most widely used framework for video encoding with multimodal inputs~\cite{gabeur2020Learning,le2020multimodal,li2020bridging,sun2019videobert}. We adopt the architecture used in \cite{li2020bridging} and train the network using the same next utterance prediction loss for FUP.
Note that this architecture is slightly different from that of BERT, and hence we cannot use pre-trained BERT weights. \\
In addition to our full model, we also show results without the object level features referred to as `CoMVT (Scene feats only)' in Table~\ref{tab:how2fup}, as this is more similar to previous multimodal models~\cite{gabeur2020Learning,le2020multimodal,li2020bridging,sun2019videobert}, as well as show the effects without BERT pretraining for the text stream and the MLM loss.
For each model including the baselines, we perform grid search on learning rates and report the test performance of the best models in the validation set.
On HowToFUP, every network is trained for 2M with a batch size of 512.
The learning rate is warmed up for 10K iterations and is continuously decayed per every 30K iterations by the factor of 0.95.
On the other datasets, due to the small sizes of the datasets, the models are trained for 20K iterations with a 50 iteration warm-up period and 1K decay length.

\subsection{Results}
\subsubsection{Future Utterance Prediction}

\begin{table}[t]
    \centering
    \caption{
    Recall at $k\in\{1,5\}$ on HowToFUP. \textbf{BERT PT:} Input text and candidate encoders initialised using BERT pretrained weights.  \textbf{MLM:} Masked Language Modelling loss. $\dagger$Multimodal Single-stream baseline used in a number of works~\cite{li2020bridging,sun2019learning,sun2019videobert}.
 $S=2$ by default for all rows except the last one where $S=4$.}
    \label{tab:how2fup}
    \scalebox{0.85}{
        \begin{tabular}{lccccc}
            \toprule
                        %& Pretrained    & \\
             Model    & BERT PT  & MLM & $S=4$ & R@1 & R@5 \\ \midrule
             S3D (Vision only) & & & & 5.85 & 18.24\\ \hdashline
            \multirow{2}{*}{Baseline (Text only)} & & & & 59.21 & 81.31\\
             %TxtTRM[BERT;Fix]~\cite{devlin2018bert}& & 27.58 & 52.11\\
             & $\checkmark$ & & & 60.90 & 82.59\\  \hdashline
             Multimodal Baseline$\dagger$ & & & & 62.73 & 84.78\\ 
             \midrule
             \multirow{3}{*}{\begin{tabular}{@{\hskip 0cm}l@{\hskip 0cm}} 
             CoMVT \\(Scene feats only)\end{tabular}} & & & & 63.35 & 85.22\\
             &$\checkmark$& & & 65.64 & 86.67\\
             &$\checkmark$& $\checkmark$& & 67.13 & 87.42\\ \hdashline
             \multirow{3}{*}{\begin{tabular}{@{\hskip 0cm}l@{\hskip 0cm}} 
             CoMVT \\(Combined feats)\end{tabular}}  &$\checkmark$&  & &  66.82 & 87.38\\
             &$\checkmark$& $\checkmark$& & 67.79 & 88.00\\ 
             &$\checkmark$& $\checkmark$& $\checkmark$& \bf68.34 & \bf88.28\\ \bottomrule
        \end{tabular}
    }
\end{table}

Table~\ref{tab:how2fup} shows the recall at $k\in\{1, 5\}$ (R@$k$) on HowToFUP.
All multimodal models outperform text only baselines showing the value of visual inputs for this task. Our best model results in an 8\% improvement in R@1. The gain due to visual input is also demonstrated by the examples  in Figure~\ref{fig:qualitative} and Appendix~\ref{sec:qual}.

\begin{table}[t]
    \centering
    \caption{
    Flops and R@$k$s with and without compact feature set extraction on HowToFUP.
    All models trained with MLM. Note we cannot train with $S=4$ without compact feature set extraction due to memory constraints.
    }
    \label{tab:compact_features}
    \scalebox{0.85}{
        \begin{tabular}{lcccc}
            \toprule
            Model             & $S$   & Gflops        & R@1   & R@5 \\ \midrule
            w/o compact         & 2     & 8.5           & 67.83 & 88.05 \\ 
            feature set extraction  & 4     & 12.7          & N/A   & N/A \\ \midrule
            w/ compact          & 2     & 6.8 (-20.2\%) & 67.79 & 88.00  \\
            feature set extraction  & 4     & 7.7 (-39.3\%) & 68.34 & 88.28  \\ \bottomrule
        \end{tabular}
    }
\end{table}

\begin{figure}[t]
    \centering
    \includegraphics[width=0.85\linewidth]{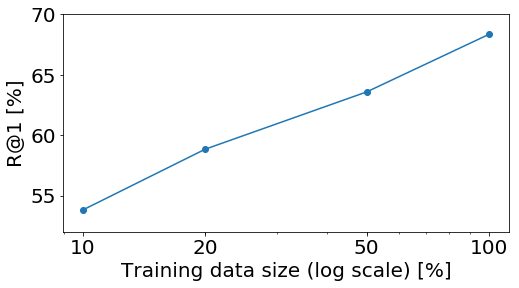}
    \caption{Effects of training data size on HowToFUP performance, reported as R@1.}
    \label{fig:dataset_size}
\end{figure}

We next ablate various aspects of our model and training setup. \\
\noindent\textbf{Architecture Components:}
%\phseo{(Effect of compact features)}
Table~\ref{tab:how2fup} shows the incremental value of different components in our architecture.
Our two stream model surpasses the single stream multimodal baseline model while using the same scene-level features.
It is also interesting that both BERT pretraining and the masked language modeling loss improve performance even though we train on a large-scale dataset with more than 35M training examples.
Finally, the use of the combined features in our full model and additional CoTRM blocks show additional gains.\\
\noindent\textbf{Efficiency:}
We also analyze the efficiency gains of our compact feature set extraction module (see Table~\ref{tab:compact_features} for flops\footnote{Flops are measured per sample by profiling evaluation steps on TPUs.} and R@$k$ results). 
Using our compact feature set module significantly reduces flops by 20.2\% and 39.3\% with $S=2$ and $4$, respectively, while maintaining performance with $S=2$.
For $S=4$, we are unable to obtain R@$k$s without compact feature set extraction on our TPU configurations due to significant memory consumption during training.\\
\noindent\textbf{Effect of Training Data:}
We also perform an ablation study analysing the effect of training data size on performance. We train on different fractions of the HowToFUP training set (results in Figure~\ref{fig:dataset_size}), and show steep performance drops when the size of the training set is reduced. We also note that the trend points towards linear improvements in R@1 as the training set size is doubled. Given that the performance does not seem to be saturated yet, we hypothesise that further performance gains are possible by scaling up with instructional video data beyond HowTo100M. 

Future Utterance Prediction results for COIN are shown in Table~\ref{tab:coin_fup}. Similar trends hold, however we also note that pretraining on HowToFUP provides a massive boost in performance (over 30\% value increase in R@1). This significant gain can be explained by the relatively small size of COIN-FUP.

\begin{table}[t]
    \centering
    \caption{Results on COIN-FUP. $S=1$ by default.
    \textbf{HowTo PT}: Entire model is pretrained on HowToFUP with $S=4$. }
    \label{tab:coin_fup}
    \scalebox{0.85}{
        \begin{tabular}{lcccccc}
            \toprule
             Model & BERT PT & MLM & R@1 & R@5 \\ \midrule
             S3D (Vision only) & & & 2.97 & 10.44 \\ \hdashline
             \multirow{2}{*}{Baseline (Text only)} &&& 19.82 & 43.19 \\
             & $\checkmark$ & & 35.68 & 64.42 \\\hdashline
             Multimodal Baseline & & & 21.02 & 45.82 \\ \midrule
             \multirow{3}{*}{CoMVT} & & & 22.40 & 47.25 \\
             & $\checkmark$ & & 37.08 & 66.90 \\
             & $\checkmark$ & $\checkmark$ & 39.11 & 68.22 \\ \hdashline
             CoMVT (HowTo PT) & & $\checkmark$ & \bf70.92 & \bf93.52 \\ \bottomrule
        \end{tabular}
    }
\end{table}

\begin{figure*}[t]
    \centering
    \scalebox{0.8}{
        \begin{tabular}{cc}
            \toprule
            \textbf{Inputs (video frames and utterances)} & \textbf{Prediction (next step class)} \\ \midrule
            
            % Example1: -humHGoKXzw-00053
            \begin{tabular}{@{}p{15.5cm}@{}}
                \centering
                \adjincludegraphics[valign=M,width=0.325\linewidth]{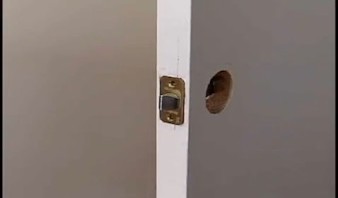}
                \adjincludegraphics[valign=M,width=0.325\linewidth]{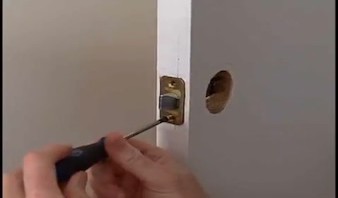}
                \adjincludegraphics[valign=M,width=0.325\linewidth]{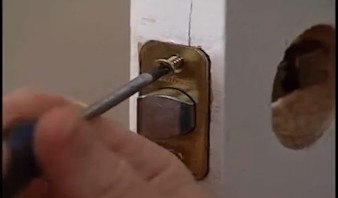} \tabularnewline[1.5cm]
                \textbf{Transcript}: The process is simply reversed insert the new \underline{dead light} into the hole at the end of the door.  
            \end{tabular}
            & 
            \bgroup
            \def\arraystretch{1.3}
            \begin{tabular}{p{3.8cm}c}
                \cellcolor[rgb]{0.7725,0.8784,0.7059}
                \raggedright \textit{install new door knob} &
                \adjincludegraphics[valign=B,width=0.025\linewidth]{figures/qualitative_examples/checkmark.png}
                \tabularnewline
                \raggedright \textit{boil water or coffee}\tabularnewline
                \cellcolor[rgb]{0.9784,0.7725,0.7059}
                \raggedright \textit{install bulb and light housing} \tabularnewline
                \raggedright \textit{close switch} \tabularnewline
                \raggedright \textit{drive car forward} \tabularnewline[2mm]
                \raggedright $\boldsymbol{\cdots}$ 730 more step classes $\boldsymbol{\cdots}$\tabularnewline
            \end{tabular}
            \egroup \\
            \bottomrule

        \end{tabular}
    }
    \caption{
    \textbf{Qualitative results on COIN-NSP.} 
    On the right, we show the predicted class (out of 735 classes) by the text only baseline (\colorbox[rgb]{0.9784,0.7725,0.7059}{red}) and the correct class identified by our multimodal model (\colorbox[rgb]{0.7725,0.8784,0.7059}{green}). 
    Note that an ASR error (highlighted by underline, `dead light' should be `deadlatch') causes the text only baseline to make a mistake (prediction about light), which is corrected by access to visual inputs in our model.
    }
    \label{fig:qualitative_nsp}
\end{figure*}

\subsubsection{Next Step Prediction}

The results for Next Step Prediction (a classification task) on COIN-NSP and CrossTask-NSP are provided in Table~\ref{tab:coin_nsp}.
Interestingly, using visual inputs shows more of an improvement over the text only baseline for this task compared to FUP. Note that NSP contains more examples of humans activities (while FUP has a more diverse set of utterances). 
In addition to the baselines described in Section~\ref{sec:baseline}, we also compare to 
two state-of-the-art vision-only models for next step prediction, RULSTM~\cite{furnari2020rolling} and TAR~\cite{sener2020temporal}\footnote{We reimplemented these models to use our extracted features to provide a fair comparison.}

Pretraining on BERT and using the MLM loss particularly help prevent our multimodal model from overfitting on this small dataset.
The largest gain comes from pretraining on HowToFUP, which also allows us to increase the value of $S$ in the model without suffering from overfitting.
We also show results of our model using visual inputs only (with pretrained weights), and find that we obtain decent results (this is achieved by feeding in a dummy text input with an empty sentence, \ie, a sequence containing a [CLS] token and a [SEP] token). 
A qualitative result is shown in Figure~\ref{fig:qualitative_nsp}.

\begin{table}[t]
    \centering
    \caption{
    Results on COIN-NSP and CrossTask-NSP. 
    *fixed dummy text is fed to utilize visual inputs only.}
    \label{tab:coin_nsp}
    \scalebox{0.85}{
        \begin{tabular}{@{\hskip 0.8mm}l@{\hskip 0mm}c@{\hskip 2mm}c@{\hskip 1mm}c@{\hskip 2mm}c@{\hskip 2mm}c@{\hskip 2mm}c@{\hskip 2mm}c@{\hskip 0.8mm}}
            \toprule
            &&&&\multicolumn{2}{c@{\hskip 5mm}}{COIN}&\multicolumn{2}{@{\hskip -0.1mm}c}{CrossTask} \\
             Model & \shortstack{HowTo\\PT} & \shortstack{BERT\\PT} & MLM & R@1 & R@5 & R@1 & R@5 \\ \midrule
             S3D (Vision only) & & & & 29.20 & 69.99 & 29.34 & 72.38 \\
            RULSTM (Vision only) & & & & 26.38 & 56.70 & 28.39 & 67.90\\
            TAR (Vision only) & & & & 18.13 & 33.64 & 17.64 & 46.03\\\hdashline
             \multirow{2}{*}{Baseline (Text only)} & & & & 16.31 & 42.80 & 17.77 & 50.59 \\
             & & $\checkmark$ & & 20.54 & 47.13 & 20.29 & 58.83 \\\hdashline
             Multimodal Baseline & & & & 24.97 & 65.26 & 30.73 & 74.09 \\ \midrule
             \multirow{4}{*}{CoMVT} & & & & 28.20 & 67.37 & 30.66 & 70.51\\
             & & $\checkmark$ & & 30.31 & 73.01 & 31.77 & 75.59\\
             & & $\checkmark$ & $\checkmark$ & 33.33 & 74.52 & 33.73 & 78.21 \\
             & $\checkmark$ & & $\checkmark$ & \bf37.46 & \bf76.64 & \bf42.12 & \bf81.45 \\\hdashline
             CoMVT (Vision only*) & $\checkmark$ & & $\checkmark$ & 33.84 & 68.78 & 39.19 & 81.26 \\   \bottomrule
        \end{tabular}
    }
\end{table}

\begin{table}[t]
    \centering
    \caption{Comparison to state-of-the-art on MSVD-QA and MSRVTT-QA. 
    We report top 1 accuracy [\%]. We show results of our model with and without pretraining on HowToFUP. }%\arshasays{Paul I think we remove MLM here, it's not adding much and is confusing} }
    \label{tab:vqa_msrvttqa}
    \scalebox{0.85}{
        \begin{tabular}{lcc}
            \toprule
             Methods & MSVD-QA & MSRVTT-QA \\ \midrule
             ST-VQA~\cite{jang2017tgif} & 31.3 & 30.9\\
             Co-Mem~\cite{gao2018motion} & 31.7 & 32.0 \\
             AMU~\cite{xu2017video} & 32.0 & 32.5 \\
             HMEMA~\cite{fan2019heterogeneous} & 33.7 & 33.0 \\ 
             HRA~\cite{chowdhury2018hierarchical} & 34.4 & 35.1 \\ 
             SSML~\cite{amrani2020noise} & 35.1 & 35.1 \\ 
             HCRN~\cite{le2020hierarchical} & 36.1 & 35.6 \\ \midrule
             %Ours (scratch) & 34.1* & 36.1 \\
             CoMVT (scratch) & 35.7 & \bf37.3 \\
             CoMVT (pretrained) & \bf42.6 & \bf39.5 \\  \bottomrule
        \end{tabular}
    }
\end{table}

\begin{table}[t]
    \centering
    \caption{
    Comparison to state-of-the-art on ActivityNet-QA.
 }
   %\arshasays{Let's remove MLM}}
    \label{tab:vqa_activitynetqa}
    \scalebox{0.85}{
        \begin{tabular}{lc}
            \toprule
             Model & Accuracy \\ \midrule
             E-VQA~\cite{yu2019activitynet} & 25.10 \\
             E-MN~\cite{yu2019activitynet} & 27.10 \\
             E-SA~\cite{yu2019activitynet} & 31.80 \\
             MAR-VQA~\cite{zhuang2020multichannel} & 34.60 \\ \midrule
             %Ours & 77.23 \\
             CoMVT (scratch)  & \bf36.63 \\
             CoMVT (pretrained)  & \bf38.75 \\  \bottomrule
        \end{tabular}
    }
\end{table}

\begin{table}[t]
    \centering
    \caption{
    Comparison to state-of-the-art on the How2QA validation set.
 }
   %\arshasays{Let's remove MLM}}
    \label{tab:vqa_how2qa}
    \scalebox{0.85}{
        \begin{tabular}{lc}
            \toprule
             Model & Accuracy \\ \midrule
             HERO~\cite{li2020hero} & 74.10 \\ \midrule
             %Ours & 77.23 \\
             CoMVT (scratch)  & \bf78.04 \\
             CoMVT (pretrained)  & \bf82.29 \\  \bottomrule
        \end{tabular}
    }
\end{table}

\subsubsection{Transfer Learning to Video QA}
We additionally show results of our model pretrained on HowToFUP and then fine-tuned on 4 popular video QA benchmarks, in Table~\ref{tab:vqa_msrvttqa} for MSRVTT-QA and MSVD-QA, Table~\ref{tab:vqa_activitynetqa} for ActivityNet-QA, and Table~\ref{tab:vqa_how2qa} for How2QA. 
For all datasets, our network architecture trained from scratch already outperforms or performs comparably to the existing state of the art; and finetuning from pretrained weights on HowToFUP provides a further boost.
We note that our model is pretrained without any QA supervision at all, and generalises well to video QA. 

% !TEX root = ../egpaper_for_review.tex

\section{Conclusion}
We propose a new visually conditioned  Future  Utterance  Prediction  (FUP)  learning task, where the goal is to predict the next utterance in an instructional video using both visual frames and transcribed speech. We set benchmarks on both the HowTo100M and COIN datasets, and show state-of-the-art results on downstream video QA benchmarks. We hope that this work will increase interest in the exciting field of visually contextualized dialogue systems.
\label{sec:conclusion}

{\small
\bibliographystyle{ieee_fullname}
\bibliography{egbib}
}

% \title{Look Before you Speak: Visually Contextualized Utterances \\ {\normalfont\itshape Supplementary Material}}
% \author{}

% \maketitle

\setcounter{section}{0}
\renewcommand{\thesection}{\Alph{section}}%
\setcounter{table}{0}
\renewcommand{\thetable}{\Alph{table}}%
\setcounter{figure}{0}
\renewcommand{\thefigure}{\Alph{figure}}%

\clearpage
\section*{Appendix}
\appendix
We describe further experiments ablating our model (Section~\ref{sec:ablate}), provide further insight into the experimental set up of the tasks used in the paper  (Section~\ref{sec:tasks_app}), display more qualitative results (Section~\ref{sec:qual}), and analyse the effect of ASR noise in HowToFUP with a brief manual study (Section~\ref{sec:asrerror}).

\section{Further Ablations on HowToFUP} \label{sec:ablate}

We ablate our model described in Section~4 of the main paper, varying (i) the number of CoTRM blocks $S$ (described in Section~4.1.3 of the main paper), (ii) the MLM loss, and (iii)~the visual input feature type (scene features only vs. combined features) in Table~\ref{tab:how2fup_ablation}.
With scene features only, and without the MLM loss, performance degrades rapidly as $S$ is increased (almost 4\% drop). 
Using combined features (object and scene) however, prevents this performance drop, as does 
adding in the MLM loss. Adding both together, gives the best performance with $S=4$, suggesting that the gains are complementary. 

\begin{table}[h]
    \centering
    \caption{
    Ablations of our network on HowToFUP. We vary the value of $S$ and show results with and without the masked language modeling (MLM) loss.}
    \label{tab:how2fup_ablation}
    \scalebox{0.85}{
        \begin{tabular}{lccccc}
            \toprule
             & & \multicolumn{2}{c}{w/o MLM} & \multicolumn{2}{c}{w/ MLM} \\
            Methods & $S$ & R@1 & R@5 & R@1 & R@5 \\ \midrule
            Scene features only & 1 & 65.43 & 86.52 & 66.73 & 87.10 \\ 
             & 2 & 65.64 & 86.67 & 67.13 & 87.42 \\ 
             & 4 & 61.05 & 83.58 & 67.15 & 87.44 \\ \midrule
            Combined features & 1 & 66.74 & 87.30 & 67.70 & 87.95 \\
             & 2 & \bf66.82 & \bf87.38 & 67.79 & 88.00  \\
             & 4 & 66.53 & 87.18 & \bf68.34  & \bf88.28  \\ \bottomrule
        \end{tabular}
    }
\end{table}

% \phseo{
% Additionally, we test our full model trained on different fractions of the training set to investigate the impact of the training dataset size.
% The results in Figure~\ref{fig:dataset_size} clearly indicates that collecting a large-scale dataset is critical for achieving high performance on FUP.
% In fact, our experiments show linear improvements of R@1 for every doubling of training dataset size.
% }
% \begin{figure}[h]
%     \centering
%     \includegraphics[width=0.8\linewidth]{figures/supplementary/dataset_size.png}
%     \caption{\phseo{Results of our full model tested with different training dataset sizes on How2FUP.}}
%     \label{fig:dataset_size}
% \end{figure}

\section{Configurations in Different Tasks} \label{sec:tasks_app}

\begin{figure*}[p]
    \centering
    \includegraphics[width=\linewidth]{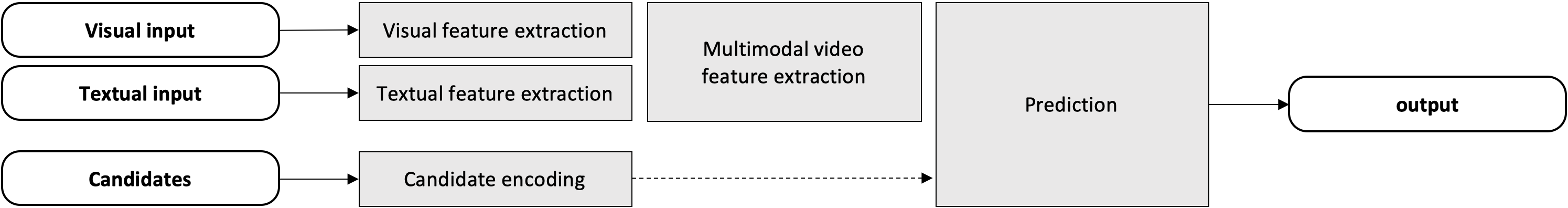} \\
    \vspace{3mm}
    \scalebox{0.85}{
        \begin{tabular}{llllll}
            \toprule
            Task & Visual Input & Textual Input & Candidates & Prediction module & Output \\ \midrule
            FUP & video frames & transcript & 100 pre-selected candidates & dot-product  & scores for utterance candidates\\
            NSP & video frames & transcript & None & two-layered classifier & scores for step classes \\
            VideoQA & video frames & transcript + question & all answers in dataset& dot-product  & scores for all answers in dataset \\
            \bottomrule
        \end{tabular}
    }
    \caption{
    Task-specific input output configurations for CoMVT. Note that for NSP, we do not use a candidate encoder since the prediction module is a classifier.
    }
    \label{fig:inputs}
\end{figure*}

In the main paper, we show results for 3 different tasks, Future Utterance Prediction, Next Step Prediction, and Video Question Answering. Here we describe the different setups for each one, in particular the inputs and outputs of our model (depicted visually in Figure~\ref{fig:inputs}). 
% Different tasks presented in our main paper have different input and output data. 
% Here, we describe how to feed the inputs to the proposed model and predict outputs in each task, which is .
\paragraph{Future utterance prediction}
In the default configuration for FUP, our model ingests video frames and transcribed speech, and expects a set of next utterance candidates.
Our model then returns a score for each candidate computed by a dot product of the input multimodal feature with each candidate.
% Computing a score for each candidate, our network returns the top scoring candidate index as the output.

\paragraph{Next step prediction}
Since this task is formulated as a classification task, instead of encoding a set of FUP candidates, we use a two-layered classifier for prediction. Inputs are video frames and transcribed speech, and output is a softmax 735-way classifier prediction.
Note that the new classifier module in this task cannot be initialized with the pretrained weights on HowToFUP and is trained from scratch.

\paragraph{Video question answering}
The goal here is to answer a question given an input video.
Compared to the other tasks, there is an input question in addition to video frames and corresponding transcribed speech.
We simply concatenate this additional input question to the transcript and feed the concatenated string as a single textual input to our model.
Note that some videos do not contain any speech and, in such cases, the model simply takes the question only.

It is common to formulate VideoQA as a classification task using the most frequent answers as target classes.
We instead adopt the formulation of answer ranking as in FUB where all possible answers in the training set are encoded using the candidate encoder and scored by the softmax normalized dot-product.
In other words, we extract all possible answers from the training set, treat them as candidate answers, and select the best candidate using the same type of a candidate encoder as in FUP.

\section{Further Qualitative Results} \label{sec:qual}
\begin{figure*}[p]
    \centering
    \scalebox{0.8}{
        \begin{tabular}{cc}
            \toprule
            \textbf{Inputs (video frames and utterances)} & \textbf{Prediction (future utterance)} \\ \midrule
            
            % Example1: 01ymdyv3umE-00010
            %   YT id: 01ymdyv3umE
            %   start_time: 53.329s
            %   end_time:   61.559s
            \begin{tabular}{@{}p{13.3cm}@{}}
                \centering
                \adjincludegraphics[valign=M,width=0.325\linewidth]{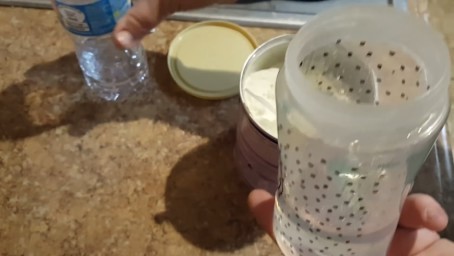}
                \adjincludegraphics[valign=M,width=0.325\linewidth]{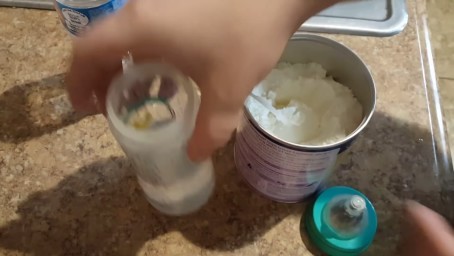}
                \adjincludegraphics[valign=M,width=0.325\linewidth]{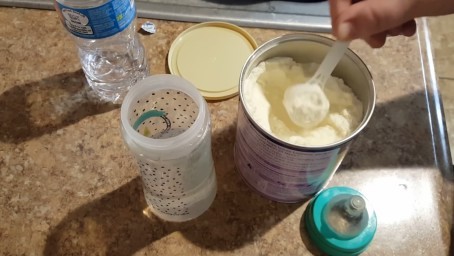} \tabularnewline[1.15cm]
                \textbf{Transcript}: This should get him through the night if we're lucky. Usually he sleeps of the night.  
            \end{tabular}
            & 
            \bgroup
            \def\arraystretch{1.3}
            \begin{tabular}{@{}p{6.5cm}c@{}}
                \cellcolor[rgb]{0.9784,0.7725,0.7059}
                \raggedright \textit{We'll see.} \tabularnewline
                \raggedright \textit{So I pushed it forward and I kind of I had to five freehand.}\tabularnewline
                \cellcolor[rgb]{0.7725,0.8784,0.7059}
                \raggedright \textit{So we need three scoops of the \textbf{formula}.} &
                \adjincludegraphics[valign=B,width=0.025\linewidth]{figures/qualitative_examples/checkmark.png}
                \tabularnewline
                \raggedright \textit{I just thought it was interesting that we have some volunteer corn coming up.}\tabularnewline
                \raggedright $\boldsymbol{\cdots}$\tabularnewline
            \end{tabular}
            \egroup \\
            \midrule

            % Example1: 036R_7agR9Q-00013
            %   YT id: 036R_7agR9Q
            %   start_time: 110.810s
            %   end_time:   117.690s
            \begin{tabular}{@{}p{13.3cm}@{}}
                \centering
                \adjincludegraphics[valign=M,width=0.325\linewidth]{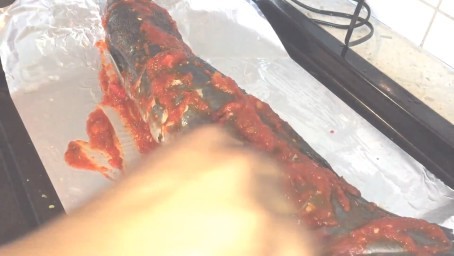}
                \adjincludegraphics[valign=M,width=0.325\linewidth]{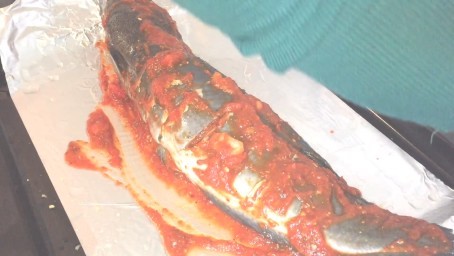}
                \adjincludegraphics[valign=M,width=0.325\linewidth]{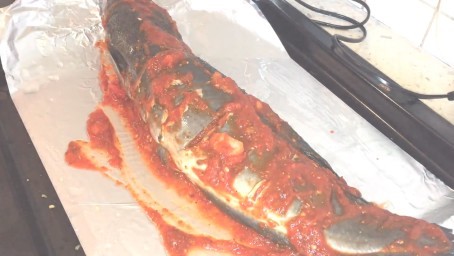} \tabularnewline[1.15cm]
                \textbf{Transcript}: Everyone was asking me why have you not been to him this stead of stuff? I don't know what else came in to me yesterday. 
            \end{tabular}
            & 
            \bgroup
            \def\arraystretch{1.3}
            \begin{tabular}{@{}p{6.5cm}c@{}}
                \raggedright \textit{This is a jewelry making tool.}\tabularnewline
                \raggedright \textit{I keep the seam gauge in my toolbox.}\tabularnewline
                \cellcolor[rgb]{0.7725,0.8784,0.7059}
                \raggedright \textit{I saw the \textbf{fish} in my fridge, and I thought what should I do with this \textbf{fish} now, and I remember Wow in Nigeria.} &
                \adjincludegraphics[valign=T,width=0.025\linewidth]{figures/qualitative_examples/checkmark.png}
                \tabularnewline
                \cellcolor[rgb]{0.9784,0.7725,0.7059}
                \raggedright \textit{I said hello just Curtis or as well be should be sue and I put in another ticket as it stated.} \tabularnewline
                \raggedright $\boldsymbol{\cdots}$\tabularnewline
            \end{tabular}
            \egroup \\
            \midrule

            % Example1: -es8DGnje6w-00004
            %   YT id: -es8DGnje6w
            %   start_time: 26.250s
            %   end_time:   39.420s
            \begin{tabular}{@{}p{13.3cm}@{}}
                \centering
                \adjincludegraphics[valign=M,width=0.325\linewidth]{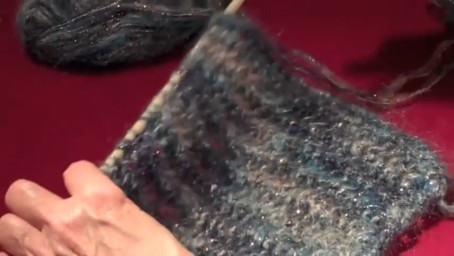}
                \adjincludegraphics[valign=M,width=0.325\linewidth]{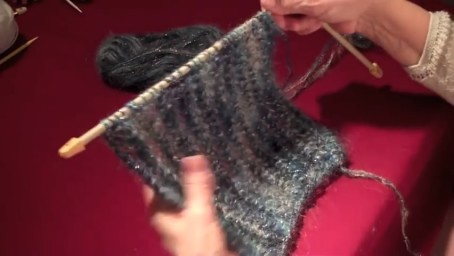}
                \adjincludegraphics[valign=M,width=0.325\linewidth]{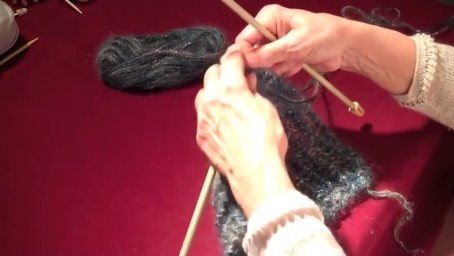} \tabularnewline[1.15cm]
                \textbf{Transcript}: So I decided to make a small scar for my cow and you know, one of those things that you put around your neck not a move on just the one that goes around your neck.
            \end{tabular}
            & 
            \bgroup
            \def\arraystretch{1.3}
            \begin{tabular}{@{}p{6.5cm}c@{}}
                \raggedright \textit{I am super full right now, but this is also good.}\tabularnewline
                \cellcolor[rgb]{0.7725,0.8784,0.7059}
                \raggedright \textit{So you're going to \textbf{knit the first stitch} bring your yarn to the front as to make a yarn over and you \textbf{knit} the next two together.} &
                \adjincludegraphics[valign=T,width=0.025\linewidth]{figures/qualitative_examples/checkmark.png}
                \tabularnewline
                \cellcolor[rgb]{0.9784,0.7725,0.7059}
                \raggedright \textit{Now the reason why I chose snakeskin, it's because I've read I've been doing a lot of research on it, but I read that snakes can't have a lot of properties for the skin.} \tabularnewline
                \raggedright \textit{He got to this stick to that bag to this stick.}\tabularnewline
                \raggedright $\boldsymbol{\cdots}$\tabularnewline
            \end{tabular}
            \egroup \\
            \midrule
            
            % Example1: 07xi-stuExU-00025
            %   YT id: 07xi-stuExU
            %   start_time: 122.870s
            %   end_time:   132.000s
            \begin{tabular}{@{}p{13.3cm}@{}}
                \centering
                \adjincludegraphics[valign=M,width=0.325\linewidth]{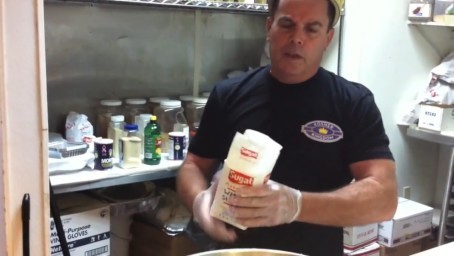}
                \adjincludegraphics[valign=M,width=0.325\linewidth]{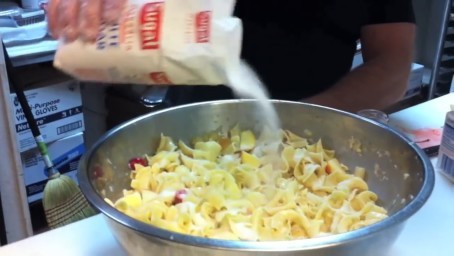}
                \adjincludegraphics[valign=M,width=0.325\linewidth]{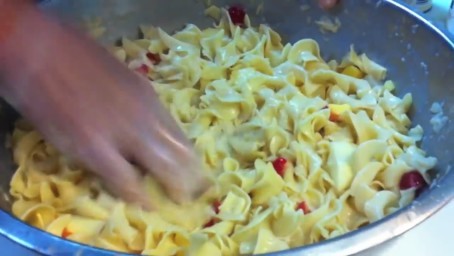} \tabularnewline[1.15cm]
                \textbf{Transcript}: It's the best sugar then you just that'll put some sugar in there to sweeten it up. That's a beautiful way to start your day.
            \end{tabular}
            & 
            \bgroup
            \def\arraystretch{1.3}
            \begin{tabular}{@{}p{6.5cm}c@{}}
                \cellcolor[rgb]{0.7725,0.8784,0.7059}
                \raggedright \textit{You want to make it diet use sweet and low beautiful \textbf{noodle kugel}.} &
                \adjincludegraphics[valign=M,width=0.025\linewidth]{figures/qualitative_examples/checkmark.png}
                \tabularnewline
                \raggedright \textit{They were like, oh it was a lot of fun.}\tabularnewline
                \cellcolor[rgb]{0.9784,0.7725,0.7059}
                \raggedright \textit{Um, the only reason why I want to do this weight loss drink today, I get the request for it.} \tabularnewline
                \raggedright \textit{You can finish eating buddy.}\tabularnewline
                \raggedright $\boldsymbol{\cdots}$\tabularnewline
            \end{tabular}
            \egroup \\
            \bottomrule

        \end{tabular}
    }
    \caption{
    \textbf{Qualitative results on HowToFUP.} 
    On the right, we show the results of the baseline model that uses text inputs only (highlighted in \colorbox[rgb]{0.9784,0.7725,0.7059}{red}) and our multimodal model (highlighted in \colorbox[rgb]{0.7725,0.8784,0.7059}{green}). The GT utterance has a $\checkmark$ next to it. 
    In many of these cases, the correct future utterance refers to an object which can only be known from the visual context (highlighted in bold). Note how both the speech and the visual frames provide complementary information, that we cannot learn from a single modality alone, helping to paint a complete picture. The ASR has mistakes (`scar' in row 3 should refer to `scarf').
    }
    \label{fig:qualitative_1}
\end{figure*}

\begin{figure*}[p]
    \centering
    \scalebox{0.8}{
        \begin{tabular}{cc}
            \toprule
            \textbf{Inputs (video frames and utterances)} & \textbf{Prediction (future utterance)} \\ \midrule
            
            % Example1: -OeaVL6-sBg-00005
            %   YT id: -OeaVL6-sBg
            %   start_time: 52.710s
            %   end_time:   59.309s
            \begin{tabular}{@{}p{13.3cm}@{}}
                \centering
                \adjincludegraphics[valign=M,width=0.325\linewidth]{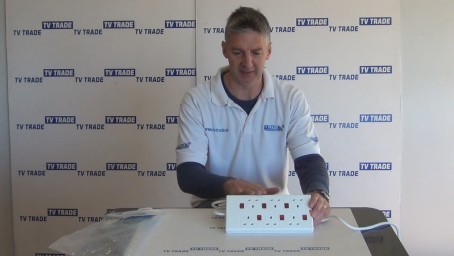}
                \adjincludegraphics[valign=M,width=0.325\linewidth]{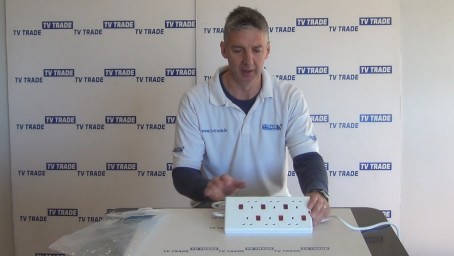}
                \adjincludegraphics[valign=M,width=0.325\linewidth]{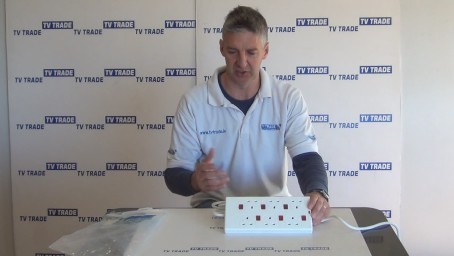} \tabularnewline[1.15cm]
                \textbf{Transcript}: The fact is, you know, got who capacity on the turkey map and the isolated scriptures on it just means it's quite an ideal community.  
            \end{tabular}
            & 
            \bgroup
            \def\arraystretch{1.3}
            \begin{tabular}{@{}p{6.5cm}c@{}}
                \cellcolor[rgb]{0.9784,0.7725,0.7059}
                \raggedright \textit{They have a tipple chassé ten minutes will be I said, it's three people three hungry people to hungry people and poor little people.} \tabularnewline
                \raggedright \textit{So I thought we would give her a try this week.}\tabularnewline
                \cellcolor[rgb]{0.7725,0.8784,0.7059}
                \raggedright \textit{So that's it anyway an overview of the \textbf{extension lead with six amperes}.} &
                \adjincludegraphics[valign=M,width=0.025\linewidth]{figures/qualitative_examples/checkmark.png}
                \tabularnewline
                \raggedright \textit{Now, what I'm doing is I've got this section of four inch pipe right here.}\tabularnewline
                \raggedright $\boldsymbol{\cdots}$\tabularnewline
            \end{tabular}
            \egroup \\
            \midrule

            % Example1: -gFKqg_UZzk-00002
            %   YT id: -gFKqg_UZzk
            %   start_time: 14.160s
            %   end_time:   26.670s
            \begin{tabular}{@{}p{13.3cm}@{}}
                \centering
                \adjincludegraphics[valign=M,width=0.325\linewidth]{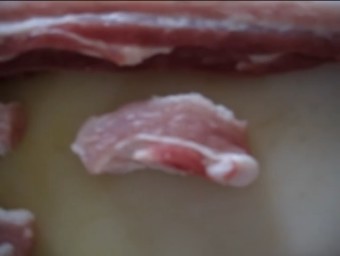}
                \adjincludegraphics[valign=M,width=0.325\linewidth]{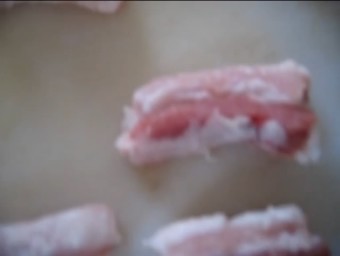}
                \adjincludegraphics[valign=M,width=0.325\linewidth]{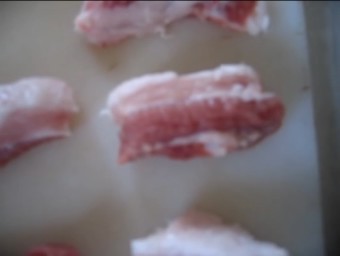} \tabularnewline[1.15cm]
                \textbf{Transcript}: You can cut into a queue about half inch thickness and say about one and half inch long but doesn't matter was the shape. Sometimes they're too big for your leave or too small for your lip.
            \end{tabular}
            & 
            \bgroup
            \def\arraystretch{1.3}
            \begin{tabular}{@{}p{6.5cm}c@{}}
                \raggedright \textit{It comes with caps.}\tabularnewline
                \raggedright \textit{So it was that kind of a medium-high.}\tabularnewline
                \cellcolor[rgb]{0.7725,0.8784,0.7059}
                \raggedright \textit{You can add more \textbf{meat}.} &
                \adjincludegraphics[valign=B,width=0.025\linewidth]{figures/qualitative_examples/checkmark.png}
                \tabularnewline
                \cellcolor[rgb]{0.9784,0.7725,0.7059}
                \raggedright \textit{You're going to continue to pick these and for gonna further a further like decorating.} \tabularnewline
                \raggedright $\boldsymbol{\cdots}$\tabularnewline
            \end{tabular}
            \egroup \\
            \midrule

            % Example1: 0kB6T84nGFE-00105
            %   YT id: 0kB6T84nGFE
            %   start_time: 435.860s
            %   end_time:   442.009s
            \begin{tabular}{@{}p{13.3cm}@{}}
                \centering
                \adjincludegraphics[valign=M,width=0.325\linewidth]{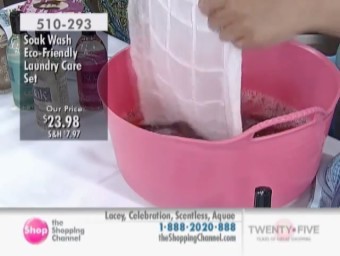}
                \adjincludegraphics[valign=M,width=0.325\linewidth]{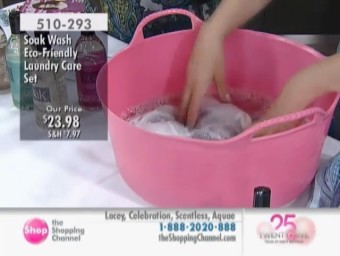}
                \adjincludegraphics[valign=M,width=0.325\linewidth]{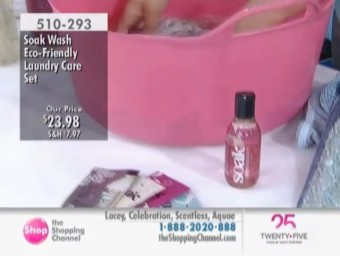} \tabularnewline[1.15cm]
                \textbf{Transcript}: It is 5 1 0 to 9 3 nice and light and fresh in terms of the fragrance, which is so lovely..
            \end{tabular}
            & 
            \bgroup
            \def\arraystretch{1.3}
            \begin{tabular}{@{}p{6.5cm}c@{}}
                \raggedright \textit{I'll show you how to change it by an industrial scene machine.}\tabularnewline
                \cellcolor[rgb]{0.7725,0.8784,0.7059}
                \raggedright \textit{You're actually going to love \textbf{washing those delicates} again.} &
                \adjincludegraphics[valign=M,width=0.025\linewidth]{figures/qualitative_examples/checkmark.png}
                \tabularnewline
                \cellcolor[rgb]{0.9784,0.7725,0.7059}
                \raggedright \textit{I mean if I was, you know, not in a hurry or whatever and I was the scent spray it on let it set for I actually tend to one time.} \tabularnewline
                \raggedright \textit{It's a violent death.}\tabularnewline
                \raggedright $\boldsymbol{\cdots}$\tabularnewline
            \end{tabular}
            \egroup \\
            \midrule
            
            % Example1: -tqKJE4v-nE-00012
            %   YT id: -tqKJE4v-nE
            %   start_time: 125.060s
            %   end_time:   130.860s
            \begin{tabular}{@{}p{13.3cm}@{}}
                \centering
                \adjincludegraphics[valign=M,width=0.325\linewidth]{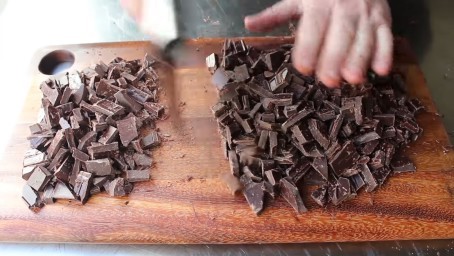}
                \adjincludegraphics[valign=M,width=0.325\linewidth]{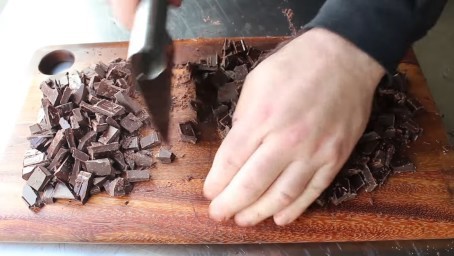}
                \adjincludegraphics[valign=M,width=0.325\linewidth]{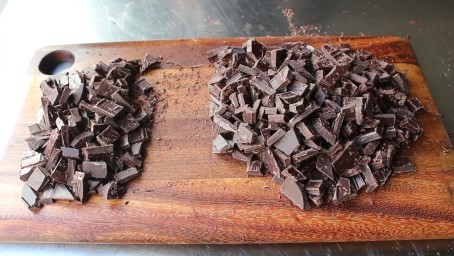} \tabularnewline[1.15cm]
                \textbf{Transcript}: So instead of that I decided to try this sounds too good to be true method to hopefully achieve the same results.
            \end{tabular}
            & 
            \bgroup
            \def\arraystretch{1.3}
            \begin{tabular}{@{}p{6.5cm}c@{}}
                \cellcolor[rgb]{0.7725,0.8784,0.7059}
                \raggedright \textit{So I'm gonna go ahead and add 2/3 of our \textbf{chocolate} to the bowl and we're just gonna do this by eye and like I said, we will reserve 1/3 of the \textbf{chocolate} to add later and to melt the \textbf{chocolate} what we'...} &
                \adjincludegraphics[valign=T,width=0.025\linewidth]{figures/qualitative_examples/checkmark.png}
                \tabularnewline
                \raggedright \textit{Nine ten.}\tabularnewline
                \cellcolor[rgb]{0.9784,0.7725,0.7059}
                \raggedright \textit{So I just put it all together all it's going to be be 2.2 percent.} \tabularnewline
                \raggedright \textit{It's loose and falls.}\tabularnewline
                \raggedright $\boldsymbol{\cdots}$\tabularnewline
            \end{tabular}
            \egroup \\
            \midrule
            
            % Example1: -glaaf_dY0E-00023
            %   YT id: -glaaf_dY0E
            %   start_time: 182.340s
            %   end_time:   187.650s
            \begin{tabular}{@{}p{13.3cm}@{}}
                \centering
                \adjincludegraphics[valign=M,width=0.325\linewidth]{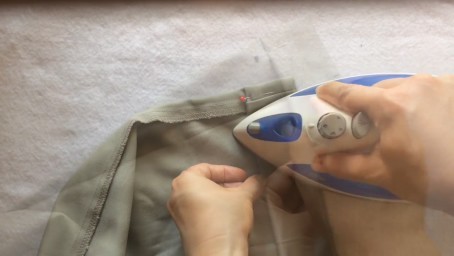}
                \adjincludegraphics[valign=M,width=0.325\linewidth]{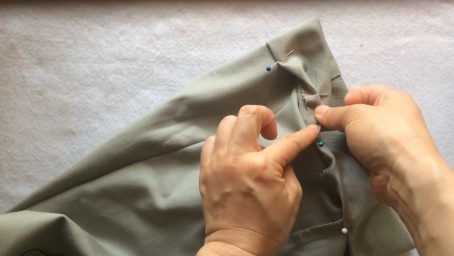}
                \adjincludegraphics[valign=M,width=0.325\linewidth]{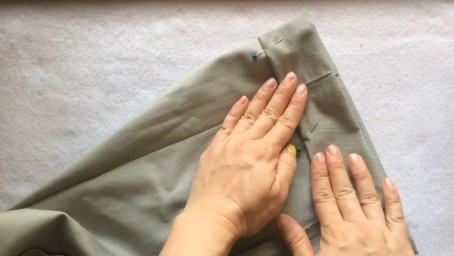} \tabularnewline[1.15cm]
                \textbf{Transcript}: So the cut ends are secured because it's tackled inside we're good.
            \end{tabular}
            & 
            \bgroup
            \def\arraystretch{1.3}
            \begin{tabular}{@{}p{6.5cm}c@{}}
                \cellcolor[rgb]{0.7725,0.8784,0.7059}
                \raggedright \textit{So let me grab my real dressed hands to study the \textbf{stitches} are around like one inch apart.} &
                \adjincludegraphics[valign=M,width=0.025\linewidth]{figures/qualitative_examples/checkmark.png}
                \tabularnewline
                \raggedright \textit{We got strawberry pina colada.}\tabularnewline
                \cellcolor[rgb]{0.9784,0.7725,0.7059}
                \raggedright \textit{So you reposition your hand on the other side to do the same thing twist up and back now we flip it on its back.} \tabularnewline
                \raggedright \textit{So that's the theory anyway, would you just put it straight in the soil?}\tabularnewline
                \raggedright $\boldsymbol{\cdots}$\tabularnewline
            \end{tabular}
            \egroup \\
            \bottomrule

        \end{tabular}
    }
    \caption{
    \textbf{Further qualitative results on HowToFUP:}     On the right, we show the results of the baseline model that uses text inputs only (highlighted in \colorbox[rgb]{0.9784,0.7725,0.7059}{red}) and our multimodal model (highlighted in \colorbox[rgb]{0.7725,0.8784,0.7059}{green}). The GT utterance has a $\checkmark$ next to it. 
    }
    \label{fig:qualitative_2}
\end{figure*}

We present additional qualitative examples for HowToFUP in Figure~\ref{fig:qualitative_1} and~\ref{fig:qualitative_2}.

\section{ASR error analysis in HowToFUP} \label{sec:asrerror}
It is a well known fact that the HowTo100M dataset is noisy, and because ASR is obtained via an automatic method, there is the potential for error.
We briefly investigate the quality of transcripts by \textit{manually correcting} ASR mistakes in the next utterances of 100 random samples from HowToFUP.
We observe a word error rate of 3.3\%, and note that at a sentence level - 80.0\% of the automatic transcripts are correct while the others contain only small (nominal) mistakes (\eg, \textit{want it} $\rightarrow$ \textit{wanted}).
To quantify the impact of these ASR errors on our model performance, we evaluate our model on the 100 samples with and without these corrected transcripts for the task of future utterance prediction. 
The model selects the identical candidates in both settings except for 2 samples (98 correct).
This indicates that the few errors introduced by ASR do not make a significant difference, particularly at scale.

% \end{multicols}
\end{document}